\pdfoutput=1

\documentclass[11pt]{article}

\usepackage[]{ACL2023}

\usepackage{booktabs}
\usepackage{bbding}
\usepackage{fdsymbol}
\usepackage{tabularx}
\usepackage{multirow}
\usepackage{amsmath}
\usepackage{epigraph}
\usepackage{amsfonts}
\usepackage{pgfplots}
\pgfplotsset{compat=1.7}
\usepackage{subcaption}
\usepackage{pifont}
\usepackage{colortbl}
\usepackage{algorithmic}
\usepackage[ruled,vlined]{algorithm2e}
\usepackage{CJK}

\usepackage{times}
\usepackage{latexsym}

\usepackage[T1]{fontenc}

\usepackage[utf8]{inputenc}

\usepackage{microtype}

\usepackage{inconsolata}

\title{Do Large Language Models Know What They Don't Know?}

\author{
Zhangyue Yin\textsuperscript{$\diamondsuit$}\quad \quad
Qiushi Sun\textsuperscript{$\spadesuit$} \quad \quad
Qipeng Guo\textsuperscript{$\diamondsuit$} \\
\bf{
Jiawen Wu\textsuperscript{$\diamondsuit$} \quad \quad
Xipeng Qiu\textsuperscript{$\diamondsuit$}\thanks{\ \ \ Corresponding author.}\quad \quad
Xuanjing Huang\textsuperscript{$\diamondsuit$}
}\\
\textsuperscript{$\diamondsuit$}School of Computer Science, Fudan University \\ 
\textsuperscript{$\spadesuit$}Department of Mathematics, National University of Singapore \\
\texttt{\{yinzy21,jwwu21\}@m.fudan.edu.cn} \quad \texttt{qiushisun@u.nus.edu} \\
\texttt{\{qpguo16,xpqiu,xjhuang\}@fudan.edu.cn}
}

\begin{document}
\maketitle
\begin{abstract}

Large language models (LLMs) have a wealth of knowledge that allows them to excel in various Natural Language Processing (NLP) tasks. Current research focuses on enhancing their performance within their existing knowledge. Despite their vast knowledge, LLMs are still limited by the amount of information they can accommodate and comprehend. Therefore, the ability to understand their own limitations on the unknows, referred to as self-knowledge, is of paramount importance. This study aims to evaluate LLMs' self-knowledge by assessing their ability to identify unanswerable or unknowable questions. We introduce an automated methodology to detect uncertainty in the responses of these models, providing a novel measure of their self-knowledge. We further introduce a unique dataset, \textit{SelfAware}, consisting of unanswerable questions from five diverse categories and their answerable counterparts. Our extensive analysis, involving 20 LLMs including GPT-3, InstructGPT, and LLaMA, discovering an intrinsic capacity for self-knowledge within these models. Moreover, we demonstrate that in-context learning and instruction tuning can further enhance this self-knowledge. Despite this promising insight, our findings also highlight a considerable gap between the capabilities of these models and human proficiency in recognizing the limits of their knowledge.

\end{abstract}
\vspace{0.2in}
\noindent ``True wisdom is knowing what you don't know.'' 
\rightline{--\textit{Confucius}}

\section{Introduction}

Recently, Large Language Models (LLMs) such as GPT-4~\citep{openai2023gpt4}, PaLM 2~\citep{anil2023palm}, and LLaMA~\citep{touvron2023llama} have shown exceptional performance on a wide range of NLP tasks, including common sense reasoning~\citep{wei2022chain,zhou2022least} and mathematical problem-solving~\citep{lewkowycz2022solving,chen2022program}. Despite their ability to learn from huge amounts of data, LLMs still have limitations in their capacity to retain and understand information. To ensure responsible usage, it is crucial for LLMs to have the capability of recognizing their limitations and conveying uncertainty when responding to unanswerable or unknowable questions. This acknowledgment of limitations, also known as ``\textit{knowing what you don't know},'' is a crucial aspect in determining their practical applicability. In this work, we refer to this ability as model self-knowledge.

\begin{figure}[t]
    \centering
 \includegraphics[width=1\linewidth]{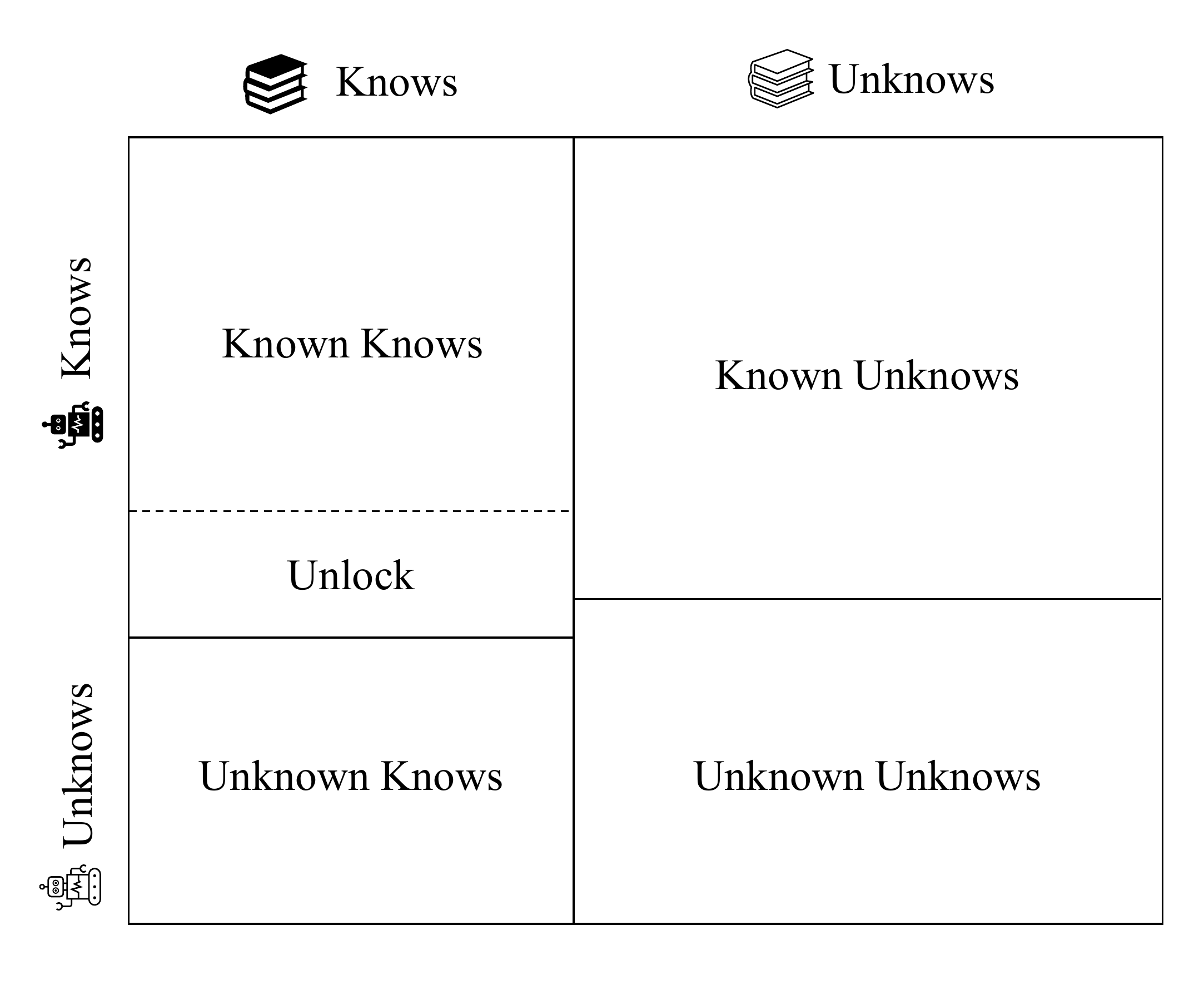}
    \vspace{-5mm}
    \caption{Know-Unknow Quadrant. The horizontal axis represents the model's memory capacity for knowledge, and the vertical axis represents the model's ability to comprehend and utilize knowledge.}
    \label{fig:Know-Unknow Quadrant}
\end{figure}

The Know-Unknow quadrant in Figure~\ref{fig:Know-Unknow Quadrant} illustrates the relationship between the model's knowledge and comprehension. The ratio of ``Known Knows'' to ``Unknown Knows'' demonstrates the model's proficiency in understanding and applying existing knowledge. Techniques such as Chain-of-Thought~\cite{wei2022chain}, Self-Consistency~\citep{wang2022self}, and Complex CoT~\citep{fu2022complexity} can be utilized to increase this ratio, resulting in improved performance on NLP tasks. We focus on the ratio of ``Known Unknows'' to ``Unknown Unknows'', which indicates the model's self-knowledge level, specifically understanding its own limitations and deficiencies in the unknows.

Existing datasets such as SQuAD2.0~\citep{rajpurkar2018know} and NewsQA~\citep{trischler2017newsqa}, widely used in question answering (QA), have been utilized to test the self-knowledge of models with unanswerable questions. However, these questions are context-specific and could become answerable when supplemented with additional information. \citet{srivastava2022beyond} attempted to address this by evaluating LLMs' competence in delineating their knowledge boundaries, employing a set of 23 pairs of answerable and unanswerable multiple-choice questions. They discovered that these models' performance barely surpassed that of random guessing. \citet{kadavath2022language} suggested probing the self-knowledge of LLMs through the implementation of a distinct "Value Head". Yet, this approach may encounter difficulties when applied across varied domains or tasks due to task-specific training. Consequently, we redirect our focus to the inherent abilities of LLMs, and pose the pivotal question: ``\textit{Do large language models know what they don’t know?}''.

In this study, we investigate the self-knowledge of LLMs using a novel approach. 
By gathering reference sentences with uncertain meanings, we can determine whether the model's responses reflect uncertainty using a text similarity algorithm. We quantified the model's self-knowledge using the F1 score. To address the small and idiosyncratic limitations of existing datasets, we created a new dataset called \textit{SelfAware}. 
This dataset comprises 1,032 unanswerable questions, which are distributed across five distinct categories, along with an additional 2,337 questions that are classified as answerable.
Experimental results on GPT-3, InstructGPT, LLaMA, and other LLMs demonstrate that in-context learning and instruction tuning can effectively enhance the self-knowledge of LLMs. However, the self-knowledge exhibited by the current state-of-the-art model, GPT-4, measures at 75.47\%, signifying a notable disparity when contrasted with human self-knowledge, which is rated at 84.93\%.

Our key contributions to this field are summarized as follows:
\begin{itemize}
\item We have developed a new dataset, \textit{SelfAware}, that comprises a diverse range of commonly posed unanswerable questions.

\item We propose an innovative evaluation technique based on text similarity to quantify the degree of uncertainty inherent in model outputs.

\item Through our detailed analysis of 20 LLMs, benchmarked against human self-knowledge, we identified a significant disparity between the most advanced LLMs and humans
\footnote{The code pertinent to our study can be accessed \href{https://github.com/yinzhangyue/SelfAware}{https://github.com/yinzhangyue/SelfAware}}.
\end{itemize}

\section{Dataset Construction}

\begin{table*}[t]
\centering
\begin{tabular}{|cccc|}
\hline
\textbf{Category}                                                 & \textbf{Description}                                                                                                                       & \textbf{Example}                                                                                                                                                                                  & \textbf{Percentage} \\ \hline
\begin{tabular}[c]{@{}c@{}}No scientific \\ consensus\end{tabular} & \begin{tabular}[c]{@{}c@{}}The answer is still up \\ for debate, with no consensus \\ in scientific community.\end{tabular}            & \begin{tabular}[c]{@{}c@{}}“Are we alone in the universe, \\ or will we discover alien \\ life at some point?”\end{tabular}                                                              & 25\%             \\ \hline

Imagination                                                            & \begin{tabular}[c]{@{}c@{}}The question are about people's \\ imaginations of the future.\end{tabular}                          & \begin{tabular}[c]{@{}c@{}}"What will the fastest form of \\ transportation be in 2050?"\end{tabular}                                                                      & 15\%             \\ \hline

\begin{tabular}[c]{@{}c@{}}Completely \\ subjective\end{tabular}   & \begin{tabular}[c]{@{}c@{}}The answer depends on \\ personal preference.\end{tabular}                                                      & \begin{tabular}[c]{@{}c@{}}"Would you rather be shot \\ into space or explore the \\ deepest depths of the sea?"\end{tabular}                                                   & 27\%             \\ \hline
\begin{tabular}[c]{@{}c@{}}Too many \\ variables\end{tabular}      & \begin{tabular}[c]{@{}c@{}}The question with too \\ many variables cannot \\ be answered accurately.\end{tabular}                            & \begin{tabular}[c]{@{}c@{}}“John made 6 dollars mowing lawns \\ and 18 dollars weed eating.\\ If he only spent 3 or 5 dollar a week, \\ how long would the money last him?”\end{tabular} & 10\%              \\ \hline

Philosophical                                                      & \begin{tabular}[c]{@{}c@{}}The question can yield \\ multiple responses, but it \\ lacks a definitive answer.\end{tabular} & \begin{tabular}[c]{@{}c@{}}“How come god was \\ born from nothingness?”\end{tabular}                                                                                                     & 23\%              \\ \hline
\end{tabular}
    \caption{Unanswerable questions in the \textit{SelfAware} dataset that span across multiple categories.}
    \label{tab:dataset}
\end{table*}

To conduct a more comprehensive evaluation of the model's self-knowledge, we constructed a dataset that includes a larger number and more diverse types of unanswerable questions than Know-Unknowns dataset~\citep{srivastava2022beyond}. To facilitate this, we collected a corpus of 2,858 unanswerable questions, sourced from online platforms like Quora and HowStuffWorks. These questions were meticulously evaluated by three seasoned annotation analysts, each operating independently. The analysts were permitted to leverage external resources, such as search engines. To ensure the validity of our dataset, we retained only the questions that all three analysts concurred were unanswerable. This rigorous process yielded a finalized collection of 1,032 unanswerable questions.

In pursuit of a comprehensive evaluation, we opted for answerable questions drawn from three datasets: SQuAD~\citep{rajpurkar2016squad}, HotpotQA~\citep{yang2018hotpotqa}, and TriviaQA~\citep{joshi2017triviaqa}. Our selection was guided by SimCSE~\citep{gao2021simcse}, which allowed us to identify and select the answerable questions semantically closest to the unanswerable ones. From these sources, we accordingly drew samples of 1,487, 182, and 668 questions respectively, amassing a total of 2,337. Given that these questions can be effectively addressed using information available on Wikipedia, the foundational corpus for the training of current LLMs, it is plausible to infer that the model possesses the requisite knowledge to generate accurate responses to these questions.

Our dataset, christened \textit{SelfAware}, incorporates 1,032 unanswerable and 2,337 answerable questions. To reflect real-world distribution, our dataset contains a proportion of answerable questions that is twice as large as the volume of unanswerable ones. Nevertheless, to ensure the feasibility of testing, we have purposefully capped the number of answerable questions.

\subsection{Dataset Analysis}

To gain insight into the reasons precluding a certain answer, we undertook a manual analysis of 100 randomly selected unanswerable questions. As tabulated in Table~\ref{tab:dataset}, we have broadly segregated these questions into five distinctive categories. ``No Scientific Consensus" encapsulates questions that ignite ongoing debates within the scientific community, such as those concerning the universe's origin. ``Imagination" includes questions involving speculative future scenarios, like envisaged events over the next 50 years. ``Completely Subjective" comprises questions that are inherently personal, where answers depend heavily on individual predispositions. ``Too Many Variables" pertains to mathematical problems that become unsolvable owing to the overwhelming prevalence of variables. Lastly, ``Philosophical" represents questions of a profound, often metaphysical, nature that resist concrete answers. Ideally, upon encountering such questions, the model should express uncertainty instead of delivering conclusive responses.

\section{Evaluation Method}
This section elucidates the methodology employed for assessing self-knowledge in the generated text. In order to achieve this, we define a similarity function, $f_{sim}$, to compute the similarity, $\mathcal{S}$, between a given sentence, $t$, and a collection of reference sentences, $U=\{u_1,u_2,\dots,u_n\}$, endowed with uncertain meanings.

\begin{equation}
\label{eq:similarity}
    \mathcal{S}_i = f_{sim}(t,u_i).
\end{equation}

Whenever any $\mathcal{S}_i$ surpasses a pre-determined threshold $\mathcal{T}$, we perceive the text $t$ as encompassing uncertain meanings, thereby eliminating the need for manual evaluation of the response.

Given the substantial disparity in the volume of answerable and unanswerable questions in \textit{SelfAware}, we adopt the F1 score as a measure of LLMs' self-knowledge. Our focus rests on identifying unanswerable questions, hence we designate them as positive cases and categorize answerable questions as negative cases.

\section{Experiment}

\begin{figure*}[t]
    \centering

 \includegraphics[width=1\linewidth]{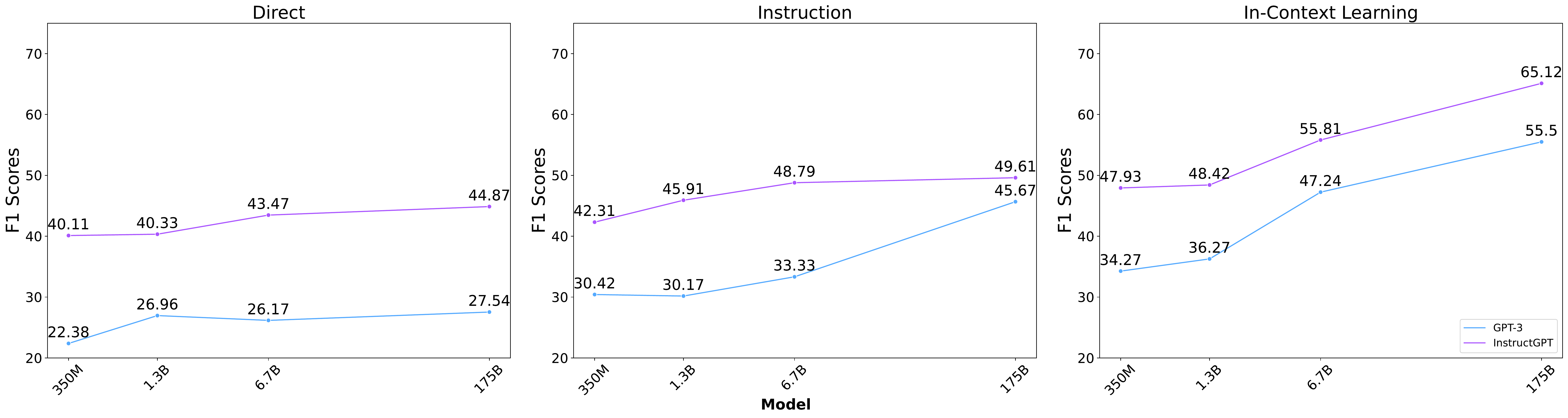}
    \vspace{-5mm}
    \caption{Experimental results using three different input forms on a series of models from GPT-3(ada, babbage, curie, and davinci) and InstructGPT(text-ada-001, text-babbage-001, text-curie-001, and text-davinci-001)}
    \label{fig:Overall}
\end{figure*}

\begin{figure}[t]
    \centering

 \includegraphics[width=1\linewidth]{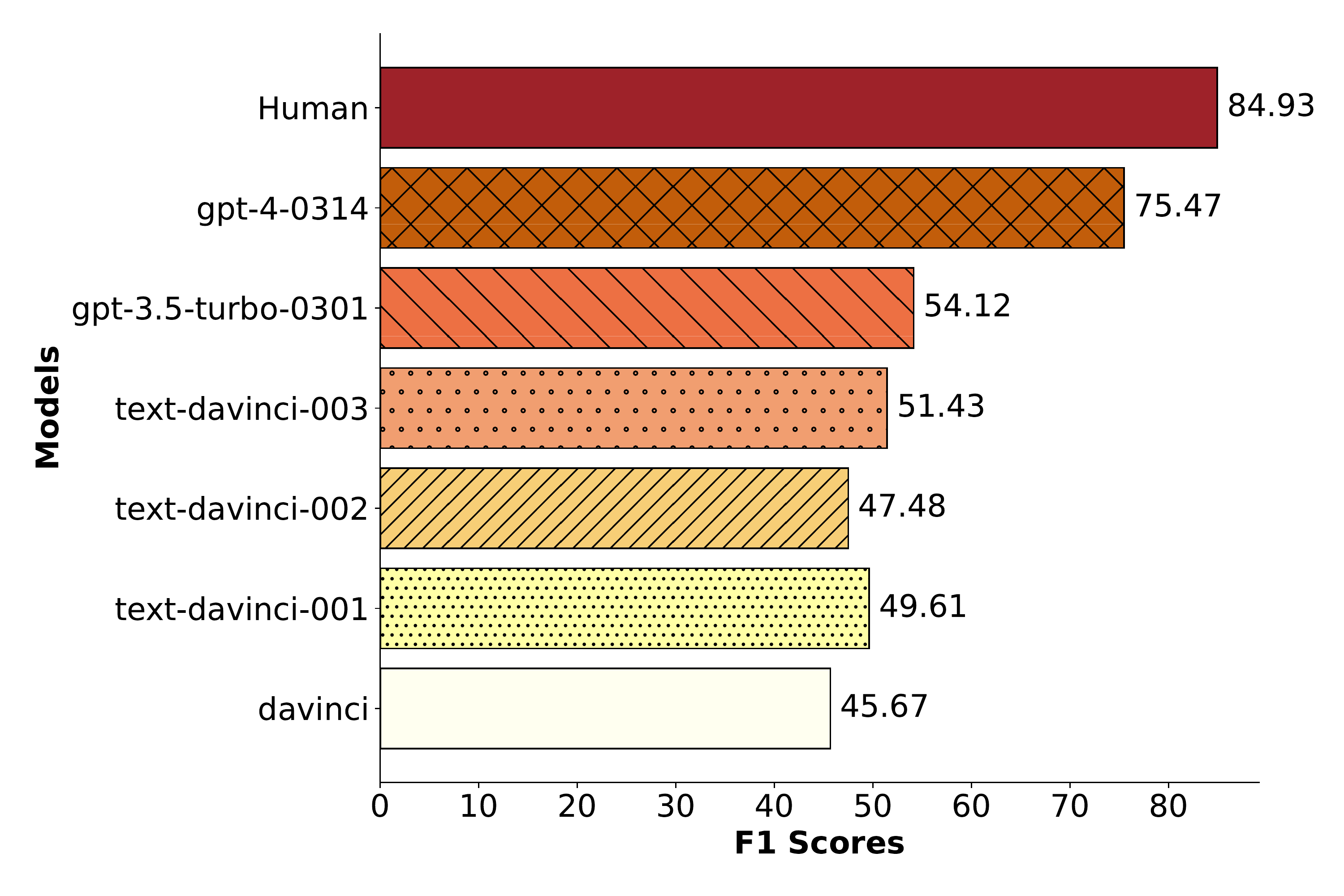}
    \vspace{-5mm}
    \caption{Comparison between the davinci series and human self-knowledge in instruction input form.}
    \label{fig:gpt-series-instruction}
\end{figure}

\begin{figure}[t]
    \centering

 \includegraphics[width=1\linewidth]{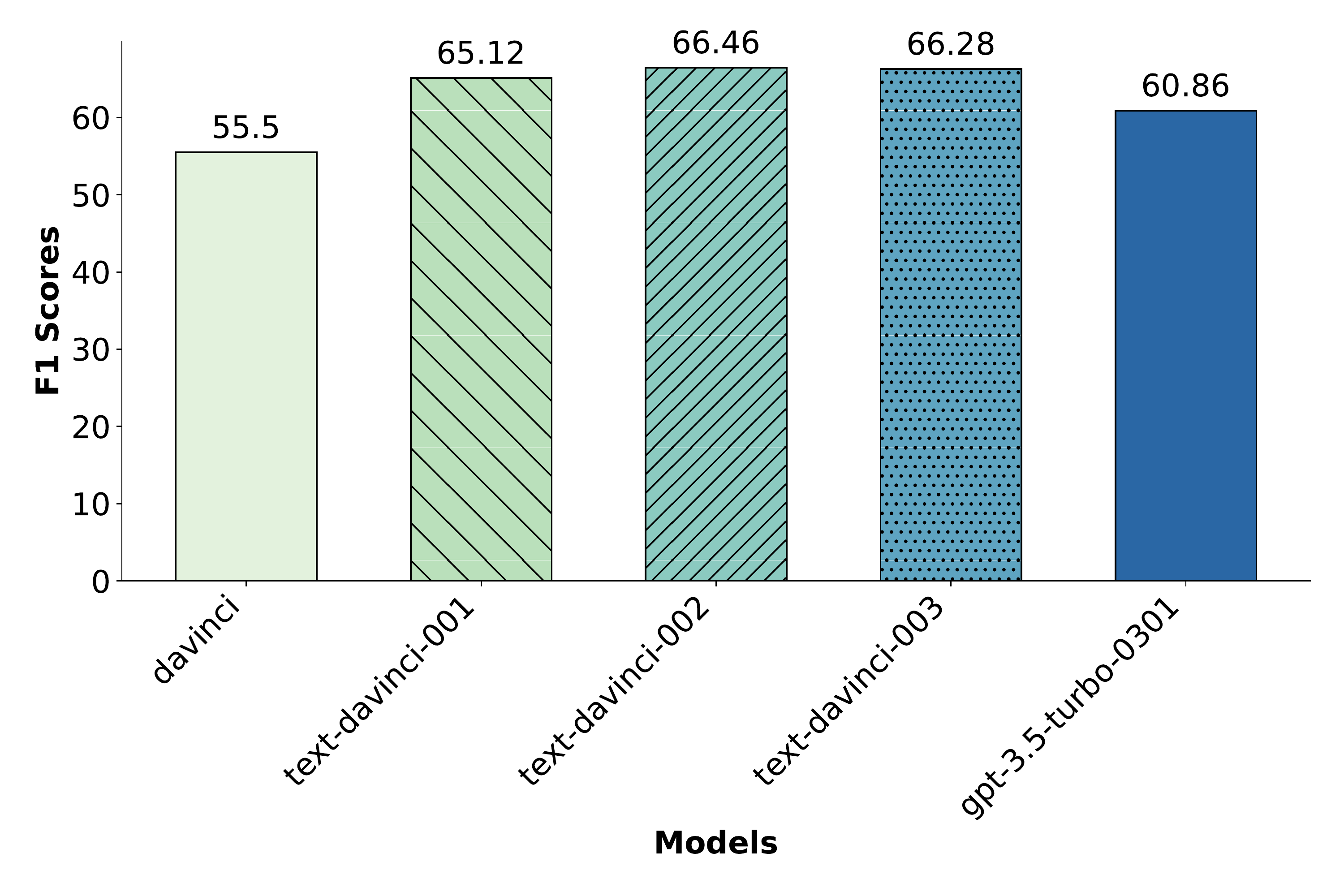}
    \vspace{-5mm}
    \caption{Experimental comparison of davinci series in ICL input form. }
    \label{fig:icl-comparsion}
\end{figure}

\subsection{Model}
We conduct a sequence of experiments to evaluate the degree of self-knowledge manifested by various LLMs, including GPT-3~\citep{brown2020language} and InstructGPT~\citep{ouyang2022training} series, as well as the recent LLaMA~\citep{touvron2023llama} and its derivative models, namely Alpaca~\citep{alpaca} and Vicuna~\citep{vicuna2023}. Our investigative approach employed three distinct input forms: Direct, Instruction, and In-Context Learning (ICL), which is encapsulated in Appendix~\ref{sec:template}.

\subsection{Setting}
We devised the reference sentence set $U$ through a process that combined automated generation by LLMs and manual filtering, detailed further in Appendix~\ref{sec:uncertainty}. To quantify the similarity between target and reference sentences, we utilized SimCSE~\citep{gao2021simcse}, setting the similarity threshold to 0.75 during our experiments. An exploration of threshold ablation is available in Appendix~\ref{sec:threshold}. To counteract potential errors in similarity calculation induced by varying lengths of the target and reference sentences, we employed a sliding window of length 5 to parse the target sentence into semantic chunks. During the generation process, we set the temperature to 0.7. We selected a random sample of 100 instances for GPT-4, while the remainder of the models were scrutinized using the full \textit{SelfAware} dataset.

\subsection{Human Self-Knowledge}
To establish a benchmark for human self-knowledge, we engaged two volunteers and selected 100 random samples from the \textit{SelfAware} dataset. The volunteers has 30 minutes to make judgments on the same set of questions, yielding an average F1 score of 84.93\%, which we subsequently adopted as the benchmark for human self-knowledge. Detailed scores are available in Appendix~\ref{sec:human_self-knowledge_test}.

\subsection{Analysis}
\begin{figure}[t]
    \centering

 \includegraphics[width=1\linewidth]{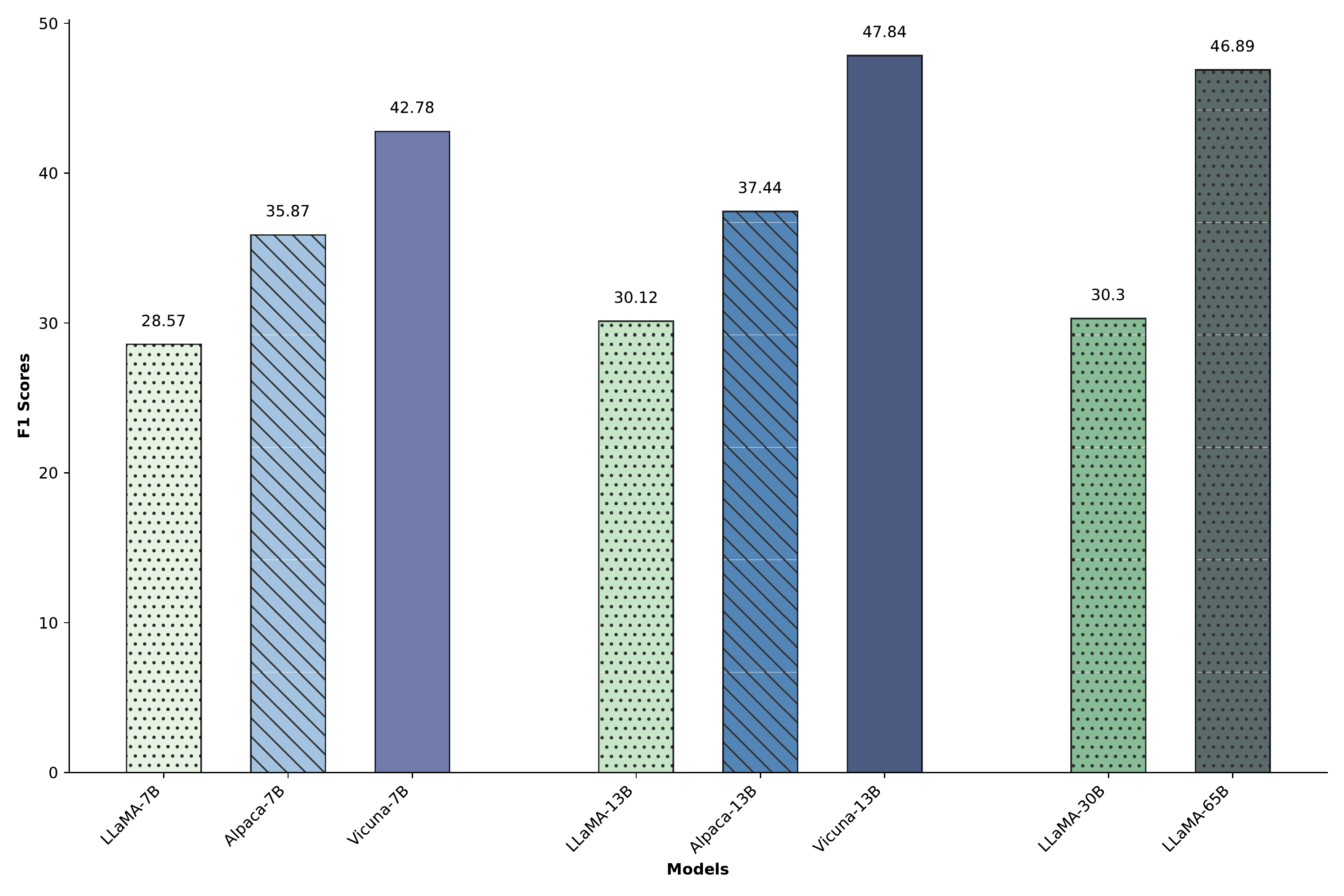}
    \vspace{-5mm}
    \caption{Experimental results obtained from LLaMA and its derived models, Alpaca and Vicuna in instruction input form.}
    \label{fig:llama-alpaca-vicuna}
\end{figure}

We evaluate the manifestation of LLMs' self-knowledge, centering our investigation on three fundamental dimensions: the size of the model, the impact of instruction tuning, and the influence exerted by different input forms.

\vspace{-.1em}
\paragraph{Model Size.} Figure~\ref{fig:Overall} illustrates the correlation between model size and self-knowledge across various LLMs. It is noteworthy that across all three input forms, an augmentation in model parameter size is associated with an elevation in the F1 Score, with the most conspicuous enhancement manifesting in the ICL input form. Therefore, our analysis indicates that an LLM's self-knowledge tends to enhance with increasing model size, a trend consistent with the scaling law.

\vspace{-.1em}
\paragraph{Instruction Tuning.} Figure~\ref{fig:Overall} delineates that models from the InstructGPT series exhibit a superior level of self-knowledge compared to their GPT-3 counterparts. Further evidence of model enhancement is provided by Figure~\ref{fig:icl-comparsion}, where text-davinci models show significant improvement relative to the base davinci model. An additional comparative analysis, presented in Figure~\ref{fig:llama-alpaca-vicuna}, evaluates LLaMA against its derivative models. The results underscore a notable increase in self-knowledge for Alpaca and Vicuna upon instruction tuning, exceeding their base model performances. Among these, Vicuna-13B outperforms the LLaMA-65B, corroborating the efficacy of instruction tuning for enhancing model self-knowledge.

\vspace{-.1em}
\paragraph{Input Forms.} As shown in Figure~\ref{fig:Overall}, the incorporation of instructions and examples serves to boost the self-knowledge of both the GPT-3 and InstructGPT series. Specifically, ICL input form, providing richer contextual information, contributes to a significant enhancement in models' self-knowledge. This impact is particularly noticeable in the davinci model, where ICL facilitates a 27.96\% improvement over the direct. Moreover, a comparison between Figure~\ref{fig:gpt-series-instruction} and Figure~\ref{fig:icl-comparsion} reveals that the inclusion of instructions and examples successfully minimizes the performance disparity between the davinci and text-davinci models, suggesting an acquisition of self-knowledge from the instructions and provided examples.

\vspace{-.1em}
\paragraph{Compared with Human.} Figure~\ref{fig:gpt-series-instruction} reveals that, without supplementary samples, GPT-4 currently performs best among the tested models, achieving an impressive F1 score of 75.47\%. However, a noticeable gap becomes evident when comparing this performance to the human benchmark of 84.93\%. This underscores the considerable potential that remains for enhancing the self-knowledge level of LLMs.

\begin{figure}[t]
    \centering

 \includegraphics[width=1\linewidth]{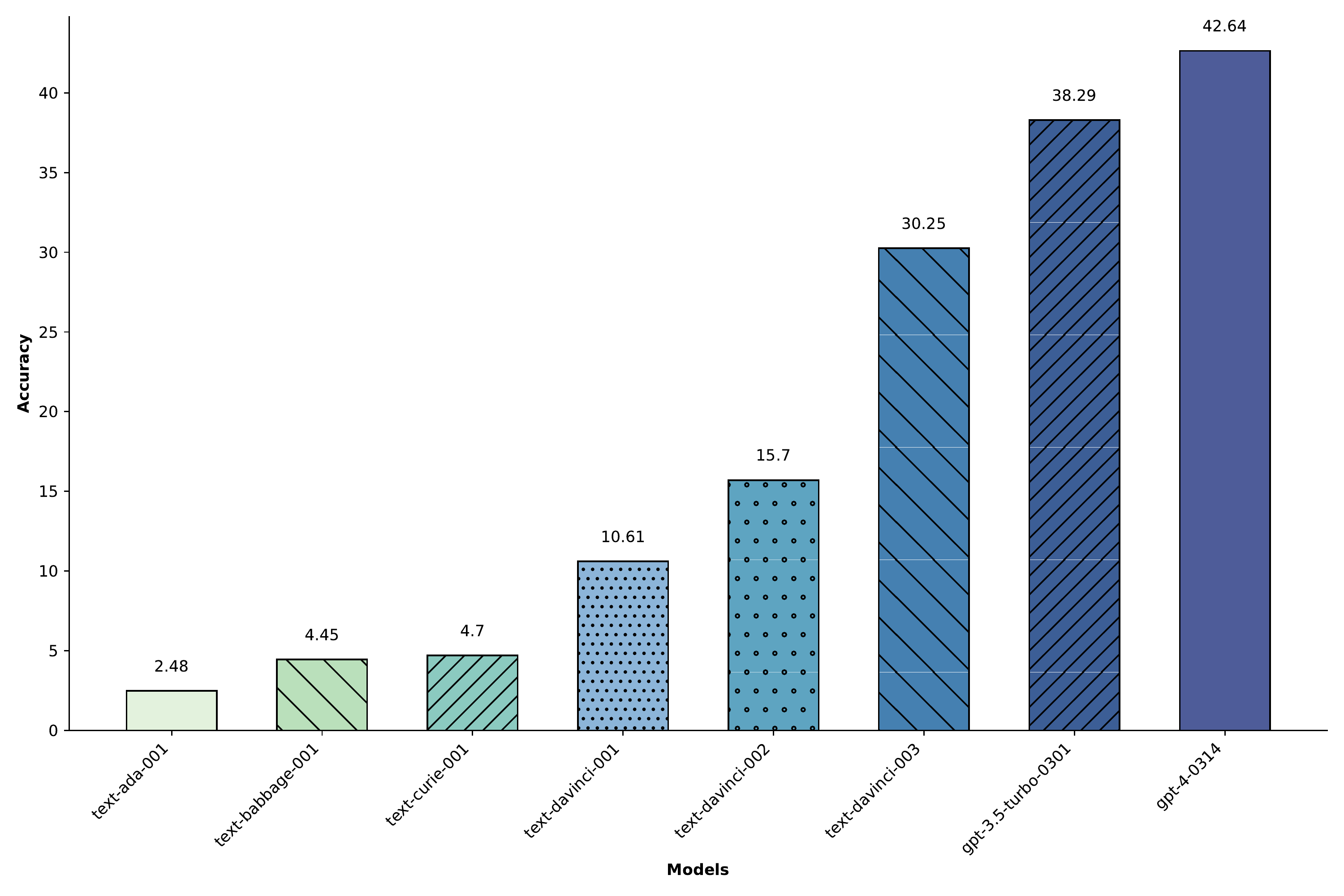}
    \vspace{-5mm}
    \caption{Accuracy of the InstructGPT series when responding to answerable questions in instruction input form.}
    \label{fig:answerable-pict}
\end{figure}

\vspace{-.1em}
\paragraph{Answerable Questions.} Figure~\ref{fig:answerable-pict} traces the performance evolution of the InstructGPT series in addressing answerable questions, adhering to the closed-book question answering paradigm~\citep{touvron2023llama}, where output accuracy is contingent on the presence of the correct answer. Our observations underscore a steady enhancement in QA task accuracy corresponding to an increase in model parameter size and continuous learning. Particularly, the accuracy of text-davinci-001 experiences a significant ascent, scaling from a meager 2.48\% in text-ada-001 to 10.61\%, whereas GPT-4 marks an even more striking jump to 42.64\%.

\section{Conclusion}
\label{sec:conclusion}

This study investigates the self-knowledge of LLMs by evaluating their ability to identify unanswerable questions. Through the introduction of a novel dataset and an automated method for detecting uncertainty in the models' responses, we are able to accurately measure the self-knowledge of LLMs such as GPT-3, InstructGPT and LLaMA. Our results reveal that while these models possess a certain degree of self-knowledge, there is still an apparent disparity in comparison to human self-knowledge. This highlights the need for further research in this area to enhance the ability of LLMs to understand their own limitations on the unknows. Such efforts will lead to more accurate and reliable responses from LLMs, which will have a positive impact on their applications in diverse fields.

\section*{Limitations}
\begin{itemize}
\item \textbf{Generalization of reference sentences.} At present, we have selected sentences with uncertain meanings exclusively from the GPT-3 and InstructGPT series, potentially overlooking uncertainty present in responses generated by other LLMs. However, it is not feasible to catalog all sentences with uncertain meanings exhaustively. As a direction for future research, we propose to concentrate on the automated acquisition of more accurate reference sentences to address this concern.

\item \textbf{Limitations of input forms:} Our examination was confined to three unique input forms: direct, instruction, and ICL. There is burgeoning research aimed at bridging the gap between models and human-like methods of reasoning and problem-solving, including but not limited to approaches like Reflexion~\citep{shinn2023reflexion}, ToT~\citep{yao2023tree}, MoT~~\citep{li2023mot}. Future endeavors will integrate additional cognitive and decision-making methods to delve deeper into the self-knowledge exhibited by these LLMs.

\end{itemize}

\section*{Ethics Statement}
The SelfAware dataset, meticulously curated to evaluate LLMs' ability to discern unanswerable questions, is composed of unanswerable questions extracted from sources such as Quora and HowStuffWorks, alongside answerable questions procured from three distinct open datasets. Every question was thoroughly examined for relevance and harmlessness. To ensure content validity, three annotation analysts, compensated at local wage standards, dedicated regular working hours to content review.

Throughout our research process, we underscored the significance of privacy, data security, and strict compliance with dataset licenses. In order to protect data integrity, we implemented anonymization and content filtration mechanisms. Our adherence to OpenAI's stipulations remained unyielding for the usage of GPT-3 and InstructGPT models, and likewise for Meta's terms pertaining to LLaMA models. We rigorously vetted the licenses of the three publicly available datasets for compliance, ensuring that all our research methodologies were in alignment with ethical standards at the institutional, national, and global levels.

Adhering to the CC-BY-SA-4.0 protocol, the dataset, once publicly released, will be reserved exclusively for research purposes. We pledge to promptly and effectively address any concerns relating to the dataset, while concurrently anticipating researchers to maintain high ethical standards in their utilization of this data.

\section*{Acknowledgement}
\label{sec:acknowledge}
We wish to express our gratitude to our colleagues in the FudanNLP group whose insightful suggestions, perspectives, and thought-provoking discussions significantly contributed to this work. Our sincere appreciation also extends to the anonymous reviewers and area chairs, whose constructive feedback was instrumental in refining the quality of our study. This work was supported by the National Natural Science Foundation of China (No. 62236004 and No. 62022027) and CAAI-Huawei MindSpore Open Fund.

\bibliography{anthology,custom}
\bibliographystyle{acl_natbib}

\clearpage

\appendix

\section{Appendix}
\subsection{Uncertainty Text}
To assemble a set of reference sentences, we randomly chose 100 entries from the \textit{SelfAware} dataset. For each model in the GPT-3 and InstructGPT series, we conducted a preliminary test using the direct input form and manually curated sentences that displayed uncertainty. From this pre-test, we procured 16 sentences manifesting uncertain connotations to serve as our reference sentences. After normalizing these sentences by eliminating punctuation and converting to lowercase, we utilized them to compute similarity with target sentences throughout our experimental procedure.

\label{sec:uncertainty}
\begin{itemize}
  \item [1.] 
  The answer is unknown.       
  \item [2.]
  The answer is uncertain. 
  \item [3.]
  The answer is unclear.
  \item [4.]
  There is no scientific evidence.
  \item [5.]
  There is no definitive answer.
  \item [6.]
  There is no right answer.
  \item [7.]
  There is much debate.
  \item [8.]
  There is no known case.
  \item [9.]
  There is no concrete answer to this question.
  \item [10.]
  There is no public information available.
  \item [11.]
  It is impossible to know.
  \item [12.]
  It is impossible to answer.
  \item [13.]
  It is difficult to predict.
  \item [14.]
  It is not known.
  \item [15.]
  We do not know.
  \item [16.]
  I'm not sure.
\end{itemize}

\subsection{Threshold ablation}
\label{sec:threshold}
We generated 100 new responses using the text-davinci-002 with direct input form and manually filtered out sentences that contained uncertainty. We then used SimCSE~\citep{gao2021simcse} to calculate the similarity between these sentences and the reference sentences in Appendix~\ref{sec:uncertainty}. We tested various thresholds for filtering sentences with uncertain meanings and compared them to manually annotated sentences. We considered unanswerable questions as positive examples and calculated precision, recall, and F1 score. The results in Table~\ref{tab:threshold} indicate that a threshold of 0.75 produced the highest F1 score, balancing precision and the inclusion of other uncertain sentences. As a result, we selected 0.75 as the similarity threshold for subsequent experiments.
\begin{table}[t]
\centering
\begin{tabular}{cccc}
\toprule
\textbf{Threshold} & \textbf{Precision} & \textbf{Recall} & \textbf{F1}    \\ 
\midrule
0.95               & 100.00             & 70.00           & 82.35          \\
0.90               & 100.00             & 75.00           & 85.71          \\
0.85               & 100.00             & 75.00           & 85.71          \\
0.80               & 100.00             & 80.00           & 88.89          \\
0.75               & 100.00             & 85.00           & \textbf{91.89} \\
0.70               & 89.47              & 90.00           & 89.73          \\
0.65               & 86.95              & 90.00           & 88.45          \\ 
\bottomrule
\end{tabular}
\caption{\label{tab:threshold}
Evaluation results comparing sentences with uncertain meaning filtered by various thresholds.}
\end{table}

\subsection{Human Self-Knowledge Test}
\label{sec:human_self-knowledge_test}
\begin{table}[t]
\centering
\begin{tabular}{cccc}
\toprule
\textbf{Human} & \textbf{Precision} & \textbf{Recall} & \textbf{F1} \\ 
\midrule
Volunteer A    & 91.52              & 78.26           & 84.37       \\
Volunteer B    & 96.36              & 76.81           & 85.48       \\ 
\bottomrule
\end{tabular}
\caption{\label{tab:human} 
Evaluation results of 100 responses from two volunteers.}
\end{table}
The evaluation results for the responses from our invited volunteers are presented in Table~\ref{tab:human}. The F1 scores for the responses were high, indicating that both volunteers exhibited a strong level of self-knowledge.

\subsection{Template}
\label{sec:template}
\begin{figure*}[t]
    \centering

 \includegraphics[width=1\linewidth]{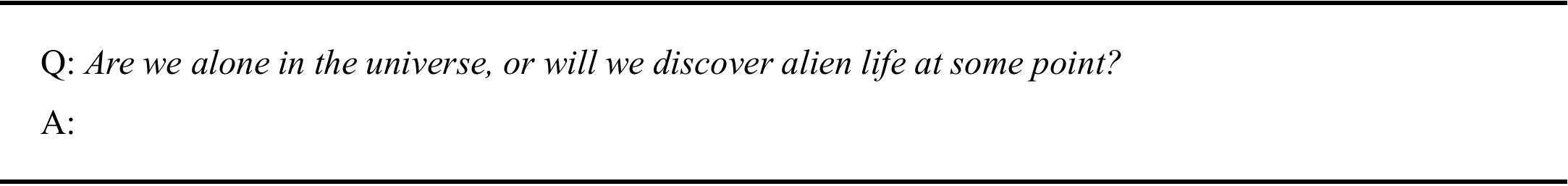}
    \caption{
    Direct input form.}
    \label{fig:direct}
\end{figure*}

\begin{figure*}[t]
    \centering

 \includegraphics[width=1\linewidth]{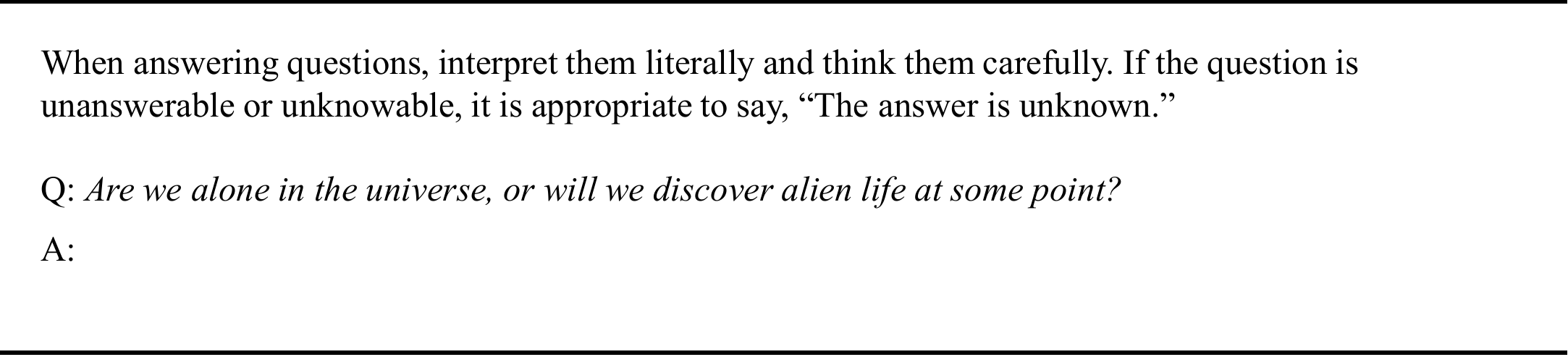}
    \caption{
    Instruction input form.}
    \label{fig:instruction}
\end{figure*}

\begin{figure*}[t]
    \centering

 \includegraphics[width=1\linewidth]{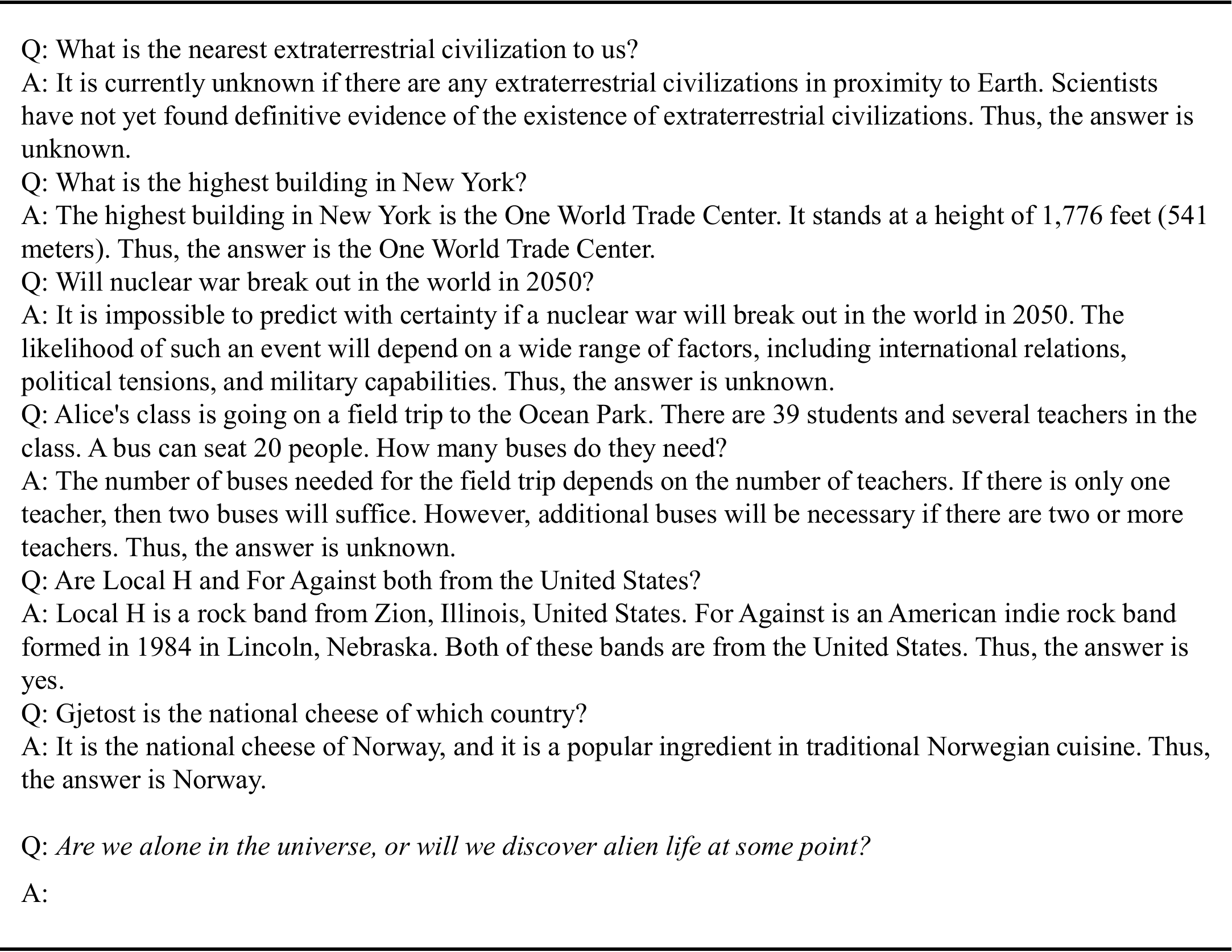}
    \caption{
    ICL input form.}
    \label{fig:icl}
\end{figure*}

The input templates used in our experiments, Direct, Instruction, and ICL, are illustrated in Figures~\ref{fig:direct}, ~\ref{fig:instruction}, and ~\ref{fig:icl}, respectively. In the ICL template, we composed 3 answerable and 3 unanswerable questions and provided the corresponding answers manually.


\end{document}